\documentclass{article}

\usepackage{arxiv}

\usepackage[utf8]{inputenc} 
\usepackage[T1]{fontenc}    
\usepackage{amsfonts}       
\usepackage{doi}
\usepackage{amssymb}

\usepackage{graphicx}                           
\usepackage{dcolumn}  
\usepackage[version=3]{mhchem}                  
\usepackage[T1]{fontenc}                        

\usepackage{epsfig}
\usepackage{epstopdf}
\usepackage{amsbsy}
\usepackage{multirow}
\usepackage{bm}
\usepackage{siunitx}
\usepackage {CJK}
\usepackage{textcomp}
\usepackage{comment}
\usepackage{nomencl}
\usepackage{parskip}                            
\makenomenclature                               

\makeatletter                                   
\renewcommand{\fnum@figure}{Figure \thefigure} 
\makeatother                                    
\usepackage[ruled,vlined]{algorithm2e}
\usepackage{mdframed}

\usepackage{color}
\definecolor{lavender}{RGB}{130,76,207}
\definecolor{teal}{RGB}{43,150,140}
\definecolor{burntumber}{RGB}{131,51,36}

\usepackage{bbm}

\newcommand{\msinstance}{\mathbf{m}}

\title{Towards Microstructural State Variables in Materials Systems}

\author{ {Veera Sundararaghavan}\thanks{Corresponding author: Prof. Sundararaghavan, Email: veeras@umich.edu, Tel: 734-615-7242} \\
	Department of Aerospace Engineering\\
	University of Michigan\\
	Ann Arbor, MI \\
	\texttt{veeras@umich.edu} \\
	\And
	{\hspace{1mm}Megna Shah and Jeff Simmons} \\
	Materials and Manufacturing Directorate\\
	Air Force Research Laboratory\\
	Wirght Patterson Air Force Base, OH \\
	\texttt{megna.shah.1@us.af.mil} \\
	\texttt{jeff.simmons.3@afrl.af.mil} \\
}

\renewcommand{\shorttitle}{\textit{arXiv} Paper}

\begin{document}
\maketitle

\begin{abstract}
The vast combination of material properties seen in nature are achieved by the complexity of the material microstructure. Advanced characterization and physics based simulation techniques have led to generation of extremely large microstructural datasets. There is a need for machine learning techniques that can manage data complexity by capturing the maximal amount of information about the microstructure using the least number of variables. This paper aims to formulate dimensionality and state variable estimation techniques focused on reducing microstructural image data. It is shown that local dimensionality estimation based on nearest neighbors tend to give consistent dimension estimates for natural images for all p-Minkowski distances. However, it is found that dimensionality estimates have a systematic error for low-bit depth microstructural images. The use of Manhattan distance to alleviate this issue is demonstrated. It is also shown that stacked autoencoders can reconstruct the generator space of high dimensional microstructural data and provide a sparse set of state variables to fully describe the variability in material microstructures. 
\end{abstract}


\newpage
\begin{mdframed}
\setlength{\nomlabelwidth}{3.2cm}
\nomenclature[01]{$p$}{Minkowski parameter}
\nomenclature[02]{$\mu$}{Intrinsic dimension}
\nomenclature[03]{$n$}{Dimensionality of ambient space of images}
\nomenclature[04]{$s$}{Number of images in the dataset}
\nomenclature[05]{$k$}{number of neighbors of a point}
\nomenclature[06]{$\bar{k}$}{Expected number of points within a distance from a point}
\nomenclature[07]{$f(\textbf{m})$}{Local probability density defining number of points per unit volume in $\Re^{\mu}$}
\nomenclature[08]{$\mathcal{V}(\mu,p)$}{volume of the p-Minkowski hypersphere of dimensionality $\mu$ that contains $\bar{k}$ nearest neighbors}
\nomenclature[09]{$V(\mu,p)$}{volume of unit hypersphere of dimensionality $\mu$}
\nomenclature[10]{$\mathcal{M}$}{random variable defining the microstructure}
\nomenclature[11]{$\mathbf{m}$}{A microstructure sample from $\mathcal{M}$, typically an image.}
\nomenclature[12]{$T_k(p)$}{p-Minkowski distance to the $k^{th}$ neighbor in the ambient space.}
\nomenclature[13]{$\bar{T}_k(p)$}{Mean over all points of distance to the $k^{th}$ neighbor in the ambient space.}
\nomenclature[14]{$r_p$}{p-Minkowski radius from a point.}
\nomenclature[15]{$c(p)$}{$f(\textbf{m})V(\mu,p)$}
\nomenclature[16]{$d_p$}{p-Minkowski distance between two points} 
\nomenclature[17]{$\lambda(p)$}{Rate of a Poisson's process}
\nomenclature[18]{$L(\mu,\theta, p)$}{log likelihood of the Poisson process}
\nomenclature[19]{$P(k-1)$}{Poisson probability of finding $k$ points in a hypersphere of radius $r_p$, with $k-1$ interior points}
\nomenclature[20]{$F_k(r_p)$}{density function ($F_k$) of a distance $r_p$ from $\textbf{m}$ to its $k^{th}$ neighbor}
\nomenclature[21]{$E_k(r_p)$}{expectation of a distance $r_p$ from $\textbf{m}$ to its $k^{th}$ neighbor}
\nomenclature[22]{$N(r_p)$}{Number of neighbors within a radius $r_p$}
\nomenclature[23]{$\theta$}{$\log f(\textbf{m})$}
\nomenclature[24]{$\mathbf{W}_i$}{weight matrix of layer i}
\nomenclature[25]{$\mathbf{b}_i$}{bias vector of layer i}
\nomenclature[26]{$\mathbf{x}$}{input vector to the encoder}
\nomenclature[27]{$\hat{\mathbf{x}}$}{output vector from the decoder}
\nomenclature[28]{$\mathbf{y}_i$}{encoded output from layer i}
\nomenclature[29]{$\sigma$}{non-linear activation function of the neural network}
\printnomenclature
\end{mdframed}

\section{Introduction}
It is widely accepted that controlling the microstructure of a material will enable control of its properties. But it is less clear which, or even how many, of the features of the microstructure represent its variability. Recently, Chen, et al. \cite{chen2022automated} identified \emph{intrinsic dimensions} in complex and chaotic dynamical systems, using only short videos of their behavior and proposed that \emph{state variables} of complex systems may be identified in this way. This suggests the tantalizing prospect of identification of a minimal set of microstructural state variables that would govern the material's behavior. This minimum number of features would encode all of the dimensions in the microstructure necessary to make design decisions, much like when the Wright brothers `invented the airplane' by discovering how to control \emph{all} dimensions of motion. Finding and controlling all dimensions of the microstructure could enable a completely new way of exploiting design spaces.

Recent advances in characterization techniques and computing have led to generation and analysis of large datasets, enabling improved understanding of microstructure. Much work has been done to quantify various aspects of the microstructure, such as particle size and shape distributions, orientation distributions, n-point statistics among others \cite{torquato2002random,ohser2001statistical,bostanabad2018computational,mssl4}, enabling significant advancements in processing-structure-properties understanding. But this has relied on domain experts manually identifying which features should be characterized. While advancing the understanding, this still leaves the uncertainty as to the degree to which the important variation in the structure was actually quantified. Although each microstructure can be represented as a vector of size $n$, the actual dimensionality of an entire database of microstructures is expected to be much lower. In formal terms, a data set containing points of dimensionality $n$ is said to have intrinsic dimensionality (ID) equal to $\mu$ < $n$ if every point lies entirely within an $\mu$-dimensional manifold of $\Re^n$. The methods of dimensionality estimation can be categorized as local and global approaches. Global methods for ID estimation rely on the spread of the entire dataset, as exemplified by projection methods such as the principal component analysis (PCA). 
Linear methods such as PCA and multidimensional scaling were explored for microstructural data in Refs \cite{thakre2021intrinsic,mssl2,mssl8}. However, it is known that such methods tend to fail on non--linear manifolds~\cite{ganapathysubramanian2008non}.  Other global approaches to dimension reduction such as ISOMAP and its variants treat non--linear manifolds using geodesic distances ~\cite{tenenbaum2000global} and have been used to reduce dimensionality of microstructures~\cite{ganapathysubramanian2008non,li2010computing}. Local approaches use the local geometry of the high dimensional space to estimate the intrinsic dimension and tend to be more computationally efficient \cite{costa2003manifold}. Levina and Bickel \cite{levina2004maximum} developed such an estimate by choosing an optimal dimension in which the local neighborhood of points would be uniformly spaced. Pope et. al. \cite{pope2021intrinsic} applied this methodology to estimate the dimensionality of some well known benchmark datasets such as the MNIST \cite{lecun1998gradient} and CIFAR \cite{krizhevsky2009learning} datasets and found that the information in those had a surprisingly low number of dimensions, i.e. degrees of freedom. Estimates ranged from 10 to 25 dimensions from the simplest to most complex dataset. 

Much of the work cited above relied on human judgement as to the reasonableness of the dimensionality estimates and did not have any ground truths by which to evaluate such reasonableness. Consequently, assessing the validity of the methods becomes problematic. This paper addresses itself to the problem of developing a self-consistent methodology for estimating the dimensionality of random media such as microstructures through validation against datasets with known dimensionality and by employing additional distance measures, all of which should yield the same estimated dimensionality. The classic work of Levina and Bickel \cite{levina2004maximum} and subsequent papers use the $L_2$ norm (Euclidean distance) to estimate the intrinsic dimension. We find that this approach is inaccurate for low bit depth images, due to the sparsity of the data. This is a systematic error that persists in recent papers, for example in Ref. \cite{chen2022automated}, where MLE estimates are higher than the intrinsic dimension for binary images. This is especially concerning for microstructure images that have a significantly reduced bit depth representing a handful of material phases. Ability to obtain consistent dimensionality estimates for generalized Minkowski distances is shown, so long as the histogram of pixel values covers a wide range, but that the estimates become inconsistent when this range is significantly reduced (known as \emph{sparsity} in the imaging literature). In this case, its is found that the $L_1$ norm, Minkowski distance for $p = 1$ is the most accurate.

ID estimators provide only the true dimensionality leaving other questions such as what state variables are actually encoded in these dimensions. More recently, deep learning generative methods have created representations that automatically capture the key variables accounting for most of the variation in image based datasets \cite{lecun2015deep,kingma2013auto}, and such models have been trained on microstructural data \cite{sundar2020database, dimiduk2018perspectives, choudhary2022recent,bostanabad2020reconstruction,fokina2020microstructure,huang2022deep}. Machine learned representations are expected to parsimoniously capture the maximal amount of information about the microstructure, as was demonstrated in Ref \cite{lubbers2017inferring} by combining neural network representation of images with manifold learning.  In  Ref. \cite{chen2022automated}, a stacked autoencoder was employed to reduce physical dynamics data to the intrinsic dimensional space.  The minimum number of variables (matching the intrinsic dimension) found from the autoencoder network are referred to as `state variables' in \cite{chen2022automated}, a terminology that is adopted in this work. To test the technique, microstructure image datasets were upsampled from a synthetic low dimensional space and passed to the stacked autoencoder. The results show a successful reduction of the images back to space describing the state variables providing a promising route to capture useful information in microstructures. 

\section{Methodology} \label{sec:method}

\subsection{Microstructures as Random Variables}

In this paper, microstructures are modelled as images whose contents are outcomes of observations of random variables~\cite{niezgoda2010stochastic}. More formally, a random variable $\mathcal{M}$ is defined to describe the \emph{Microstructure}. In this context, `Microstructure' is that used by, say, a process engineer who wants a certain microstructure because of its desirable properties. The outcomes ($\msinstance \in \Re^n$) from sampling $\mathcal{M}$ represent the images that would be observed, say, by a microscopist investigating the \emph{microstructure}, where the lower case `$\mathbf{m}$' is used to distinguish an instance from the class. Here, $n$ is the number of pixels in the image. This way, one can make use of the considerable results from sampling theory, particularly point processes\cite{snyder2012random}, in the analyses.

\subsection{Microstructures on a Manifold}

Modeling \emph{microstructure} observations as images, an image is an outcome of sampling $\mathcal{M}$ to give $\mathbf{m}$. If this image is, say $256 \times 256$ in dimension, $\mathbf{m}\in \Re^{256\times 256}$. This is a huge space, from which all images of this spatial resolution may be sampled. The vast majority of these images simply represent random noise. By hypothesis, natural (or \textit{microstructural}) images occupy a very small subset of this space. That is, the valid images that would plausibly represent a \textit{Microstructure} occupy a \emph{manifold} in $\Re^{256\times 256}$. 

\begin{figure}[ht]
    \centering
\includegraphics[width=1.0\textwidth]{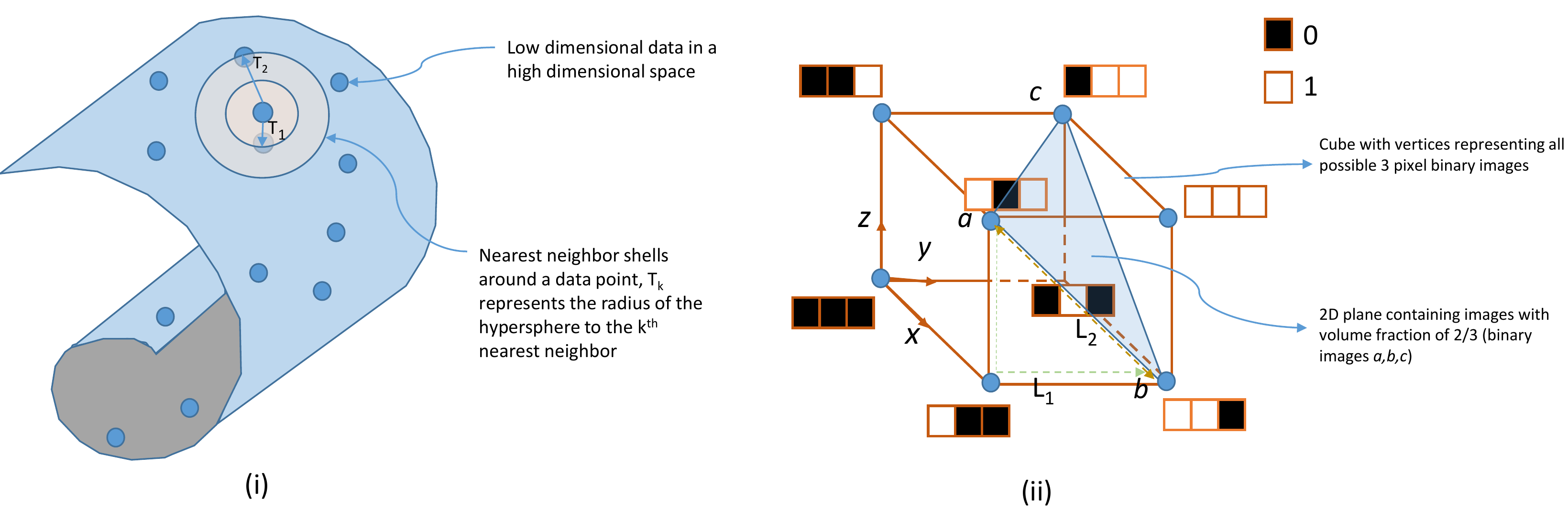}
    \caption{(i) A `swiss roll' manifold containing image data represented as points. Near--neighbor shells around a data point are illustrated which can be used to estimate the intrinsic dimensionality. (ii) A mainifold representation of binary images with n pixels, which exist on vertices of a cube of dimension n. The space of 3 pixel images are shown with a 2D domain representing images (marked a,b,c) with pixel values that sum to two.}
    \label{figschema}
\end{figure}

Speaking loosely, a manifold is a lower-dimensional space that is contained in our $\Re^{256\times 256}$ space, but having fewer total number of dimensions. A plane embedded in a 3-D space is an example of a \emph{linear manifold}, having only 2 dimensions. More generally, the term `manifold' means some non-linear subspace that can be distorted within the embedding space. Figure~\ref{figschema}(a) shows an example of a manifold in $\Re^3$, which is known as the `swiss roll' manifold. Essentially, this is a plane that contains all of the data, but has been `rolled up' into a spiral, so that it exists in $\Re^3$, but the points, themselves only occupy $\Re^2$. In this work, the manifold is referred to as a \emph{latent space} and the high--dimensional embedding space as the \emph{ambient space}. This is motivated by the fact that one would observe the images in the ambient space ($\Re^{256\times 256}$ in the current example).

Visualizing images on a latent manifold becomes problematic for greater than three dimensions: very simple images must be used for illustration, with the extension to higher dimensions being made in a more abstract sense. Using a very simple image, consisting of only 3 pixels, one can illustrate a latent manifold in an embedding space in Figure~\ref{figschema}(b). The actual ambient space is the closed set
\begin{equation}
    A = \{(x,y,z)\in \Re^3 | x\in [0,1], y\in[0,1], z\in [0,1]\}
\end{equation}
The `corners' of $A$ represent \emph{binary images}, i.e. 1-bit images, where the pixels can only have values of 0 or 1.

Within this space, a (linear) manifold is embedded as:
\begin{equation}
    B = \{(x,y,z)\in A | x + y + z = 2\}
\end{equation}
which represents binary images in which one of the pixels has a value of 0 and two have a value of 1, as well as all \emph{convex combinations} \cite{boyd2004convex} of these images to form a constrained set of grayscale images.

By hypothesis, \textit{microstructure} images occupy some latent manifold in an enormous ambient space. Obtaining this manifold is the subject of \emph{manifold learning}\cite{pless2009survey,ma2012manifold}. Our hypothesis is that the \textit{Microstructure} may be controlled by identifying state variables for its description and that these may be enumerated if one knows the dimensionality of the latent manifold on which the \textit{microstructure} images lie. It is the subject of \emph{disentanglement}, an active area of research in machine learning \cite{horan2021unsupervised,fumero2021learning}, to make these dimensions interpretable.

\subsection{Nearest Neighbor Approach to Dimensionality Estimation}

The nearest neighbor method~\cite{pettis1979intrinsic} is a geometric estimator of the intrinsic dimensionality of the manifold on which the data lies. The assumptions behind this approach are (1) that the samples are independent and identically distributed (iid) from some distribution, (2) that, in a space of proper dimension, they will be uniformly distributed, (3) that the mapping between the latent space  and the ambient space is continuous, and (4) that the distance between two points in the ambient space is the same as that in the latent space.

The intuitive meaning of `random placement,' where there is no bias towards one area in space or another. This is a common one made with modeling, say, trees in a forest. The unique point process that will assure such a random placement is the Poisson process~\cite{baddeley2007spatial}. The intuitive meaning of `continuous' is that neighboring points in the latent space correspond to neighboring points in the ambient space. Topology\cite{brown2006topology} provides a more precise statement of this, but the essential intuitive interpretation is this. 

There is one subtle complication that arises because data is generally not on a linear manifold, but on one that is curved and twisted. The distance between two points on a curved manifold would be measured as its geodesic distance, whereas, in the ambient space, it would be measured as a Euclidean distance or similar. Since differentiable manifolds are approximately Euclidean for small distances, this amounts to a requirement that the distance between points be made small.

With these assumptions, the dimensionality of a dataset may be made, knowing only a distance between the points. Levina and Bickel used the Euclidean distance, but we use the generalized $p$-norm approach of Minkowski, which reduces to the Euclidean distance for $p = 2$. All $p$-norms are required to estimate the same dimensionality, which yields a `best practice' for intrinsic dimensionality estimation.

The nearest neighbor (NN) method aims to find the intrinsic dimensionality $\mu \leq n$ using the number of nearest neighbors $k$ of each data point~\cite{levina2004maximum}. The data are modeled as being iid samples from a probability density in the low dimensional latent space $\Re^\mu$. By hypothesis, there is a locally homogeneous Poisson process, of dimension $\mu$, such that the density is constant within a neighborhood of $\mathbf{m}$ \cite{snyder2012random}, which will uniformly (at least, locally) distribute the data points in this space.

Let $\mathbf{m}_1, \mathbf{m}_2,..,\mathbf{m}_s \in \Re^n$ be the instances of $s$ \textit{microstructures}. Under these assumptions, the average number of data points ($\bar{k}$) that fall into a hypersphere in $\Re^\mu$ around a point $\textbf{m}_i$ will be proportional to the volume of the hypersphere:

\begin{equation}
    \bar{k} = f(\textbf{m}) \mathcal{V}(\mu,p) \label{eq1}
\end{equation}

where the proportionality constant, $f(\mathbf{m})$, defines the uniform probability density defining number of points per unit volume in $\Re^{\mu}$ and $\mathcal{V}(\mu,p)$ refers to the volume of the hypersphere of dimensionality $\mu$ that has an expected number $\bar{k}$ nearest neighbors with distances represented using a $L_p$ norm.

The volume of the hypersphere is given by the particular choice of the distance measure. Levina and Bickel\cite{levina2004maximum} used the Euclidean distance measure ($p = 2$) where the volume is given by the formula:

\begin{equation}
    \mathcal{V}(\mu,2) = V(\mu,2) {[T_k(2)]}^\mu  \label{eq2}
\end{equation}

Where $V(\mu,2)$  is the volume of a hypersphere of unit radius in $\Re^\mu$, $T_k(2)$ is the distance from a fixed point $\textbf{m}$ to its $k^{th}$ nearest neighbor in the ambient space, and the constant $2$ within brackets in Eq. \ref{eq2} indicates the Minkowski 2-norm, which is the Euclidean distance measure, is being used. By the locally isometric hypothesis, this is the same as the distance would be measured in the latent space. 

\subsubsection{Generalized Distance Measures}

For the Euclidean distance, the volume of a unit hypersphere is $\frac{\pi^{\mu/2}}{\Gamma(\mu/2 + 1)}$\cite{levina2004maximum}. We extend this analysis to apply to the general Minkowski distances of order $p$, ($d_p$).

\begin{figure}[ht]
    \includegraphics[width=1.0\textwidth]{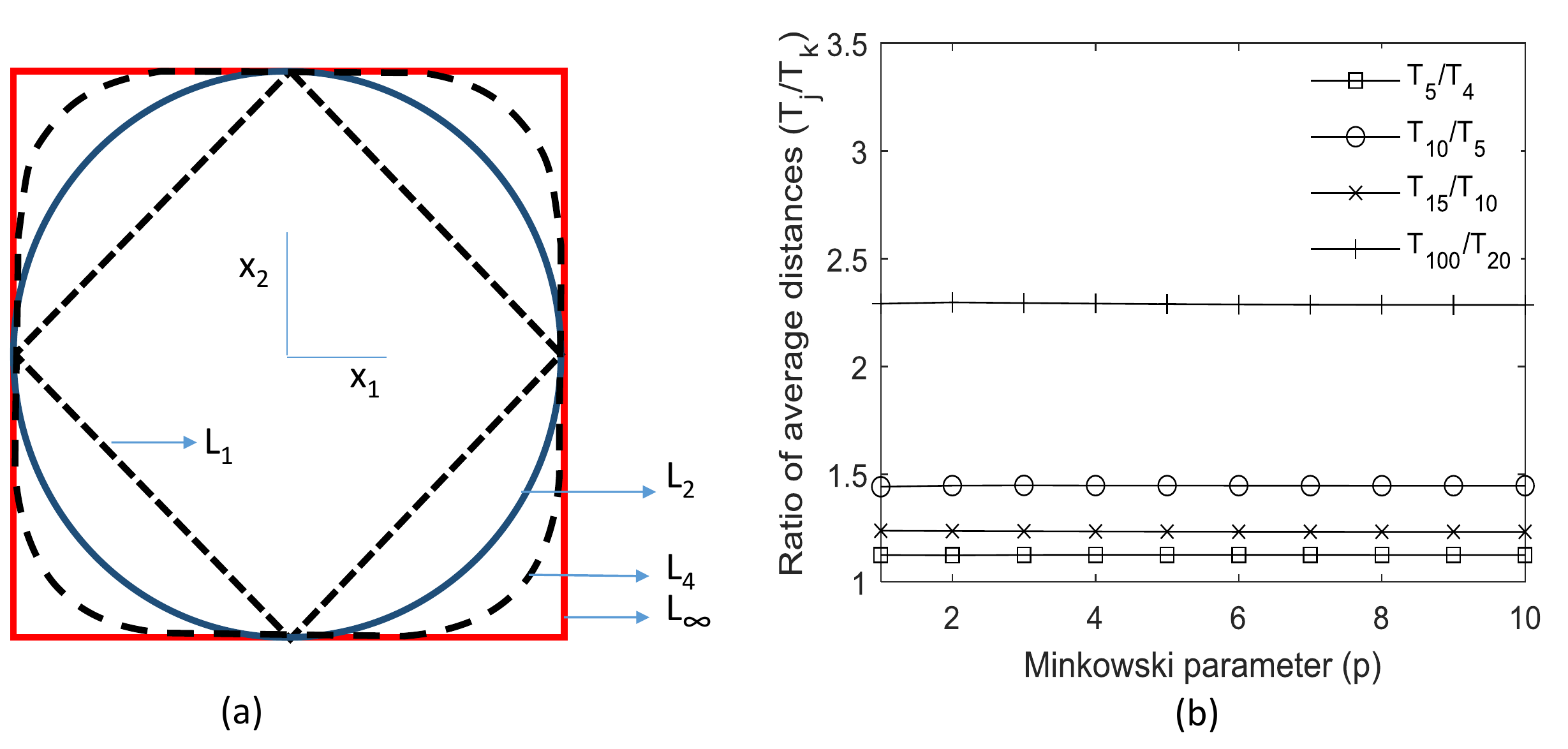}
    \caption{(a) Minkowski circles (b) Our estimation of the ratio of average distances to $j^{th}$ and $k^{th}$ nearest neighbor shells for the swiss roll dataset as a function of the Minkowski parameter.}
    \label{figmin}
\end{figure}

Between points $\textbf{m}_q$ and $\textbf{m}_l$, $d_p$ is defined as:
\begin{equation}
    d_p(\textbf{m}_q,\textbf{m}_l)  \triangleq \left(\sum_{i=1}^{n} |\textbf{m}_{q,i} - \textbf{m}_{l,i}|^p\right)^{1/p} \label{eq3}
\end{equation}
Particular cases of the Minkowski distance family are $d_1$, commonly known as the Manhattan distance or the $L_1$ norm and $d_2$, commonly known as the Euclidean distances or the $L_2$ norm. A geometric representation of a 2D \emph{circle} for $p = 1,2,4$, and $\infty$ is shown in Fig. \ref{figmin}(a), where the surface describes all points equidistant from the origin under the respective $d_p$.

The volume of a hypersphere of dimensionality $\mu$ when using a $d_p$ distance measure (see below) is:
\begin{equation}
    \mathcal{V}(\mu,p) = V(\mu,p){[T_k(p)]}^\mu  \label{eq4}
\end{equation}
where, $V(\mu,p) = \frac{2^{\mu}{[\Gamma(1/p + 1)]}^\mu}{\Gamma(\mu/p + 1)}$ is the volume of a hypersphere of unit radius in $\Re^\mu$ and $T_k(p)$ is the distance to the $k^{th}$ nearest neighbor, both being measured in terms of the $d_p$ distance.

For a choice of the Minkowski parameter $p$, the relationship in Eq. \ref{eq1} can be used to estimate the dimension by regressing $\log {\bar{T}_k(p)}$ on $\log k$ over a suitable range of $k$ (eg. from $k = k_a$ to $k = k_b$), where $\bar{T}_k(p)$ denotes the mean $d_p$ distance of points to their $k^{th}$ nearest neighbor. The intrinsic dimension is obtained as the slope:

\begin{equation}
    \mu = \frac{\log k_b - \log k_a}{\log T_{k_b}(p) - \log T_{k_a}(p)} =  \log\left(\frac{k_b}{k_a}\right) \left(\log \frac{T_{k_b}(p)}{T_{k_a}(p)}\right)^{-1}\label{eq5}
\end{equation}

Since $\mu$ is a unique intrinsic dimension, the above equation implies that the ratio $\frac{T_{k_b}(p)}{T_{k_a}(p)}$ is independent of $p$. 

This can be seen as follows, based on a Poisson point process. Using Eq. \ref{eq1} and Eq. \ref{eq4}, the expected number of points within a distance $r_p$ from a point $\textbf{m}$ can be written as:
\begin{equation}
    \bar{k} = c(p)r_p^\mu\label{eqk1}
\end{equation}

where $c(p)=f(\textbf{m})V(\mu,p)$. The hypersphere defined by the points between $\mathbf{m}$ and the $k^{th}$ neighbor contains $k-1$ points in its interior, the $k^{th}$ being on the boundary, itself. The Poisson distribution ($P$) for finding $k-1$ points within a distance of $r_p$ from point $\textbf{m}$ is given by: 
\begin{equation}
    P(k-1) = \frac{(c(p)r_p^\mu)^{k-1}}{\Gamma(k)} \exp (-c(p)r_p^\mu) \label{eqk2}
\end{equation}

From which one can infer that the rate of the Poisson process is $\lambda(p) = \frac{d}{dr} (c(p)r_p^\mu)$. 

The density function ($F_k$) of a distance $r_p$ from $\textbf{m}$ to its $k^{th}$ neighbor can be written as (using Eq.~\ref{eqk1} and \ref{eqk2}, and performing change of variables for the probability density), ~\cite{pettis1979intrinsic}:

\begin{equation}
    F_k(r_p) = \left(\frac{(c(p)r_p^\mu)^{k-1}}{\Gamma(k)} \exp (-c(p)r_p^\mu)\right)c(p)\mu r_p^{\mu-1} \label{eqk3}
\end{equation}

From this expression, the expectation of a distance $r_p$ from $\textbf{m}$ to its $k^{th}$ neighbor can be found as (see appendix 2):

\begin{equation}
    E_k(r_p) = \int_0^{\infty} r_pF_k(r_p)dr_p = c(p)^{-\frac{1}{\mu}}\frac{\Gamma(k + \frac{1}{\mu})}{\Gamma(k)} \label{eqk4}
\end{equation}

The leading term $c(p)^{-\frac{1}{\mu}}$, which is a function of Minkowski parameter p, is independent of $k$. This implies that the ratio of average distances for different values of $k$ (eg. in Eq. \ref{eq5}) will be independent of the Minkowski parameter. This is, indeed, correct, as our estimates of the ratio of average distances to $j^{th}$ and $k^{th}$ nearest neighbor shells ($\frac{T_{j}(p)}{T_{k}(p)}$) for different Minkowski parameters for the swiss roll shows, Fig. \ref{figmin}(b). 

\subsection{MLE estimation using the p--norm}
\label{SectionD}

The intrinsic dimensionality estimator in Ref. \cite{levina2004maximum} is a variant of the nearest neighbor theory which seeks a maximum likelihood estimate (MLE) instead of a mean estimate. The difference is subtle: The nearest neighbor approach estimates the dimension as a statistic that can be computed from data (as in Eq. \ref{eq5}), while the MLE approach seeks the optimum parameter in the Poisson distribution (eq. \ref{eqk2}), which in practise yields a more robust estimate of dimensionality.

The log likelihood of the Poisson process can be written as: 
\begin{equation}
L(\mu,\theta,p) = \int_0^R \log(\lambda(p)) ~dN(r_p) - \int_0^R \lambda(p)~dr_p
\end{equation}
where $N(r_p)$ is the number of points within a distance $r_p$ from $\mathbf{m}$ and $\theta = \log f(\mathbf{m})$. Maximizing the likelihood using $\frac{\partial L}{\partial \theta} = 0$ and $\frac{\partial L}{\partial \mu} = 0$, an optimal value of $\mu$ is obtained, also containing ratios of distances \cite{levina2004maximum}:

\begin{equation}
   \mu_k(\mathbf{m_i},p) = \frac{1}{k-1} \left[\sum_{j=1}^{k-1} \log\left(\frac{T_k(p)}{T_j(p)}\right)\right]^{-1}\label{eqmle}
\end{equation}

\begin{figure}[ht]
    \includegraphics[width=1.0\textwidth]{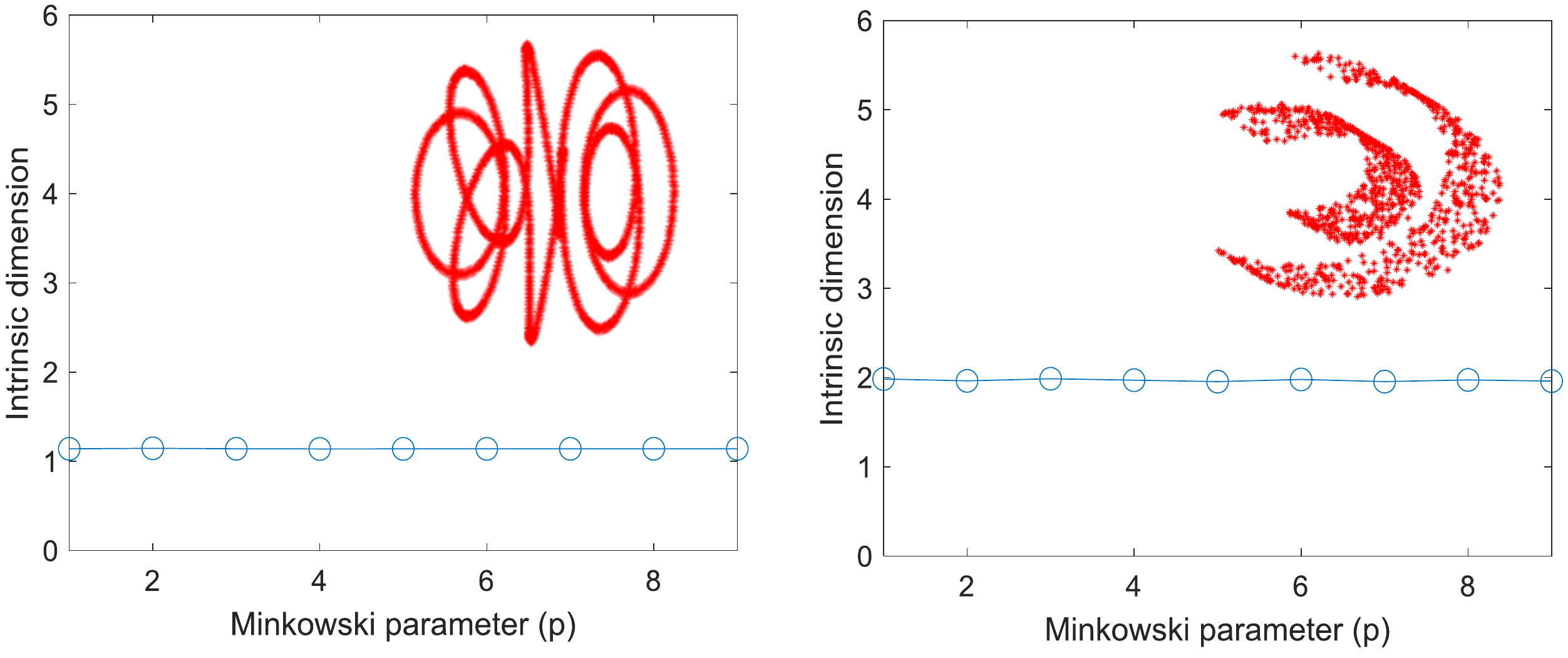}
    \caption{MLE based intrinsic dimension for different Minkowski parameters: (left) Helix, $\mu = 1$ (right) Broken swiss roll, $\mu = 2$.}
    \label{fighelix}
\end{figure}

As described in Levina and Bickel, a denominator of $k-2$ instead of $k-1$ gives an unbiased estimate and is employed in this work. Fig.~\ref{fighelix}  shows the variation of computed intrinsic dimension by this approach against the choice of Minkowski parameter for a Helix and a broken swiss roll dataset. The correct intrinsic dimension is found for all Minkowski parameters tested: 1 for the helix and 2 for the broken swiss roll. 

\subsection{Numerical considerations}

Datasets of interest in this work are material microstructures that typically consist of two phases (eg. a fiber composite, containing a fiber and a matrix) or a finite number of phases (eg. steels containing austenite, martensite, ferrite etc.). In two phase images, pixels are labeled such that the precipitate is one and the matrix is zero (binary image). Note that the independence of intrinsic dimensionality to Minkowski parameter is true only if the distance $r$ is non-discrete (eg. cases in Fig. \ref{fighelix}). Here, we show an example where this breaks down using the case of binary images. Here, one has a discrete manifold where images occupy vertices of a cube as shown in Fig. \ref{figschema}(ii). Consider the distance between any two images under the Minkowski distance family for the case of binary image. Since the absolute difference between any two pixels is either zero or one, the distance between any two image instances $\textbf{m}_q$ and $\textbf{m}_l$ can be written as:
\begin{equation}
    d_p(\textbf{m}_q,\textbf{m}_l) = (||\textbf{m}_q - \textbf{m}_l||_{1})^{1/p} \label{eqmin}
\end{equation}
Here, $||..||_{1}$ refers to the $L_1$ norm (Manhattan distance) which derives from using $\textbf{m}_{q,i} - \textbf{m}_{l,i}$ in eq. \ref{eqmin} is either zero or one for binary images. For Minkowski parameter $p = 1$, lets say that fitting $\log {\bar{T}_k(1)}$ against $\log k$ would give the slope as the intrinsic dimension $\mu^*$. When using $p=2$, substituting $T_k(2) = (T_k(1))^{1/2}$ (eq. \ref{eqmin}) would give an intrinsic dimension (slope) of $2\mu^*$ instead. The same behavior is also obtained with the MLE equation (Eq. \ref{eqmle}). In general, an intrinsic dimension of $p\mu^*$ would be obtained for the p-Minkowski measure. This inconsistency in our computed value of intrinsic dimension for binary images is related to a discretization error. One of the objectives of this paper is to identify a p-Minkowski measure that mitigates this issue in binary microstructural images (or in general, images with low bit depth) via numerical examples.

\section {Results}
\label{Results}

The results employ the MLE estimate in Eq. \ref{eqmle}. In this equation, every data point $\textbf{m}_i$ gives a dimension estimate for every neighbor count $k$. A dimension estimate matrix of size $k \times n$ is obtained where $n$ is the number of images in the dataset. The intrinsic dimension is estimated as the mean of the values in the dimension estimate matrix. An important aspect in dimensionality estimation is removing duplicates from the dataset so that the same datapoint is not double--counted as its own neighbor. This was done on all datasets before computing  the intrinsic dimension.

\subsection{Binary Datasets}

In rest of the results, the datasets are split into four categories:

\begin{itemize}
    \item Dataset 1. Rectangles and squares in a matrix. Ten different synthetic datasets were tested containing rectangular and square shapes in a matrix following different size and positional constraints which dictate the intrinsic dimensionality (shown in Fig. \ref{figcases}). Images are of size $128\times128$. In cases A, B, F, G, the shapes were randomly placed leading to two free dimensions of x and y coordinates of the center of the shape. In addition, for cases A and B varying sizes were used adding one more dimension in the case of squares (the width) and two more in the case of rectangles (width and height). In cases C, D, H and I, the shapes were placed linearly along a horizontal axis at the center, eliminating one intrinsic dimension (the y-coordinate of center) from cases A, B, F, G, respectively. The last two cases, E and J, include centered shapes with a variation only in the size of the object. In all cases, the shapes in the images are complete, ie, they are not cutoff by the image boundary.\\
    
    \item Dataset 2.  Randomly centered circles of a constant radius in a matrix. The coordinates of the center $(x,y)$ are independently sampled using uniform random variables. Images are of size $256\times256$. Four datasets 2A, 2B, 2C, 2D containing radii of $r = $ 24, 36, 48 and 58 pixels, respectively were generated with circle centers ($(x,y)$) selected within a range such that the circles do not intersect the boundaries.
    \begin{equation}
    \{(x,y)\in \Re^2| r < x < 256-r, r < y < 256-r\}
    \end{equation}
    
    \item Dataset 3. A known 3D point cloud is used as the generator for microstructural images, where each point in the 3D cloud $(x_1,x_2,x_3)$ is mapped to $(x,y,R)$ in a binary image of a circle in a matrix, where $R$ is the circle radius, $(x,y)$ is the center of the circle. In this way, the radius is no longer a free variable and is related to the center of the circle via the topology of the point cloud (termed the `generator space' to differentiate from the latent space, which could be lower in dimension). The mapping for the circle from the generator space to the ambient space would be $f:\Re^3\rightarrow \Re^{128\times 128}$.   Two cases were used \\
    (i)  Dataset 3A (swiss roll circles). The generator space is a 3D swiss roll given as:
    \begin{equation}
x = t \cos t + c_1,
y = 30\eta_2 + c_2, 
z = t \sin t + c_3
\end{equation}
where $t = \frac{3\pi}{2}(1 + 2\eta_1)$, $\eta_1$ and $\eta_2$ are uniform random variables in the range of 0 to 1 and $(c_1,c_2,c_3)$ are translation factors chosen as $(62,50,20)$ respectively. The latent space will be two dimensional, governed by the choice of the two random variables $\eta_1$ and $\eta_2$.

    (ii) Dataset 3B (helix circles). Generator space is a 3D helix, given by the equations: 
        \begin{equation}
x = 5(13+ (2 + \cos 8t) (\cos t)),
y = 5(13+ (2 + \cos 8t) (\sin t)), 
z = 9(4 + \sin 8t)
\end{equation}
where $t = 2\pi \eta$, $\eta$ being a uniform random variable in the range of 0 to 1. The latent space will be one dimensional, with the location in the manifold governed by the choice of $\eta$. Note that the coordinates of points in the generator space of Dataset 3A and 3B are rounded before mapping to images because of the integer (pixel) representation of the centers and radii. 

\item Dataset 4: Contains results from a phase field simulation of grain growth based on the Allen-Cahn equation following the numerical formulation of Fan and Chen \cite{fan1997computer}. The data is in the form of binary images ($128\times128$) containing grain boundaries at different time steps of a single simulation. Since all the model parameters are fixed at the start of the simulation and images are only a function of time, the intrinsic dimensionality of all images from a single simulation run is expected to be one. Dataset contains results from three different simulation runs.

\end{itemize}

\subsection{Intrinsic dimension estimates}

\begin{figure}[ht]
    \centering
\includegraphics[width=1.0\textwidth]{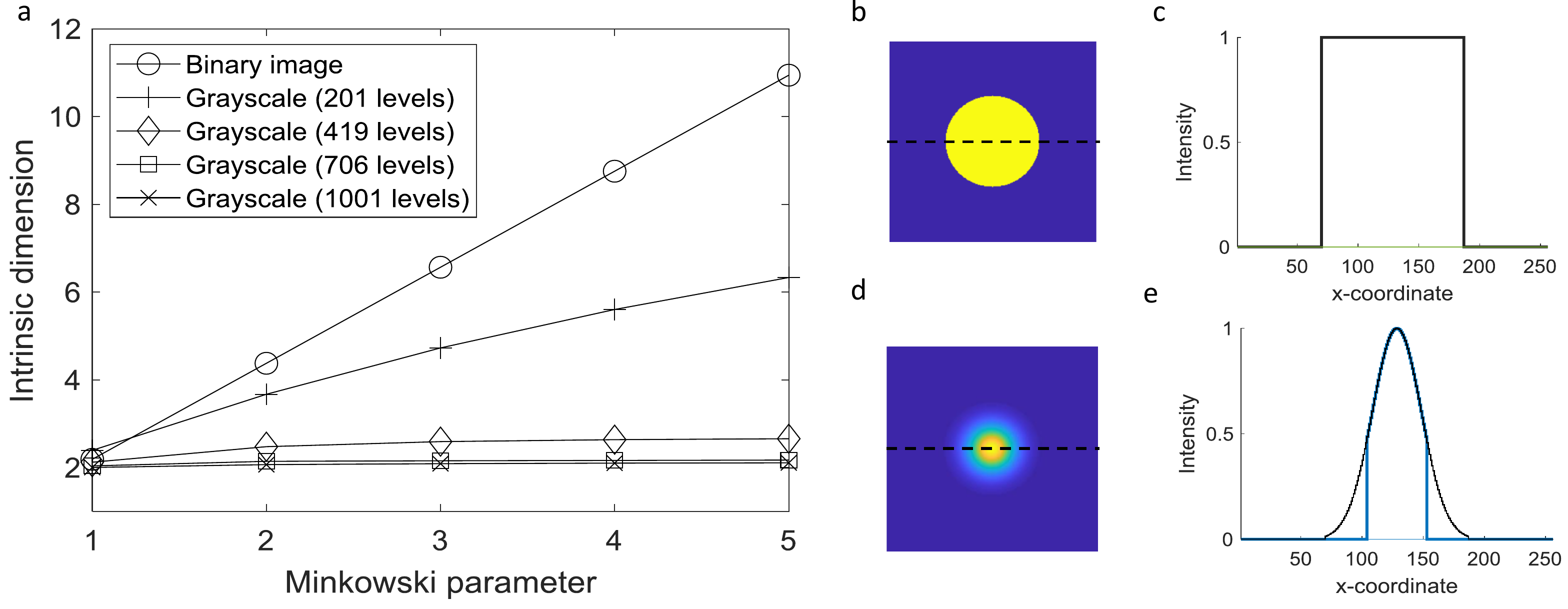}
    \caption{a. Variation of intrinsic dimension with the choice of Minkowski parameter for a circle with varying position, represented as binary and grayscale images. The true dimensionality is 2. b. Shows a binary microstructure c. Intensity across the dotted line in (b) is shown. (d) A grayscale microstructure based on a Gaussian intensity profile. As the number of grayscale levels increase, the answers for higher p-Minkowski distance measures converge toward the true estimate.}
    \label{figgauss}
\end{figure}

Fig. \ref{figgauss}(a) shows the variation of intrinsic dimension with the choice of Minkowski parameter for binary image dataset--2D. The intrinsic dimension is two, and corresponds to the  $(x,y)$ coordinate of the center of the circle. In the binary case, a linear increase in the intrinsic dimension with Minkowski parameter is obtained as explained previously. While it was expected that the $L_1$ norm gives the minimum intrinsic  dimension estimate among these cases, it is also seen that the  $L_1$ distance measure matches the expected intrinsic dimensionality. To further confirm this, the circle is replaced with a Gaussian distribution centered at the circle origin and with a constant standard deviation (of 20) for all datasets. Since each of the four cases in dataset 2 contains a circle of different radii, each case spans different number of grayscale levels. An example is shown in Fig. \ref{figgauss}(e) with the blue line spanning a part of the Gaussian curve, distance between the blue lines is the circle diameter). Cases with radii of 24, 36, 48 and 58 pixels contain 201, 419, 706 and 1001 grayscale levels, respectively. As the number of grayscale levels increase, the intrinsic dimension obtained from the use of higher p-Minkowski distance measures converge toward the true estimate as given by the $L_1$ norm. 

\begin{figure}[ht]
    \centering
    \includegraphics[width=1.0\textwidth]{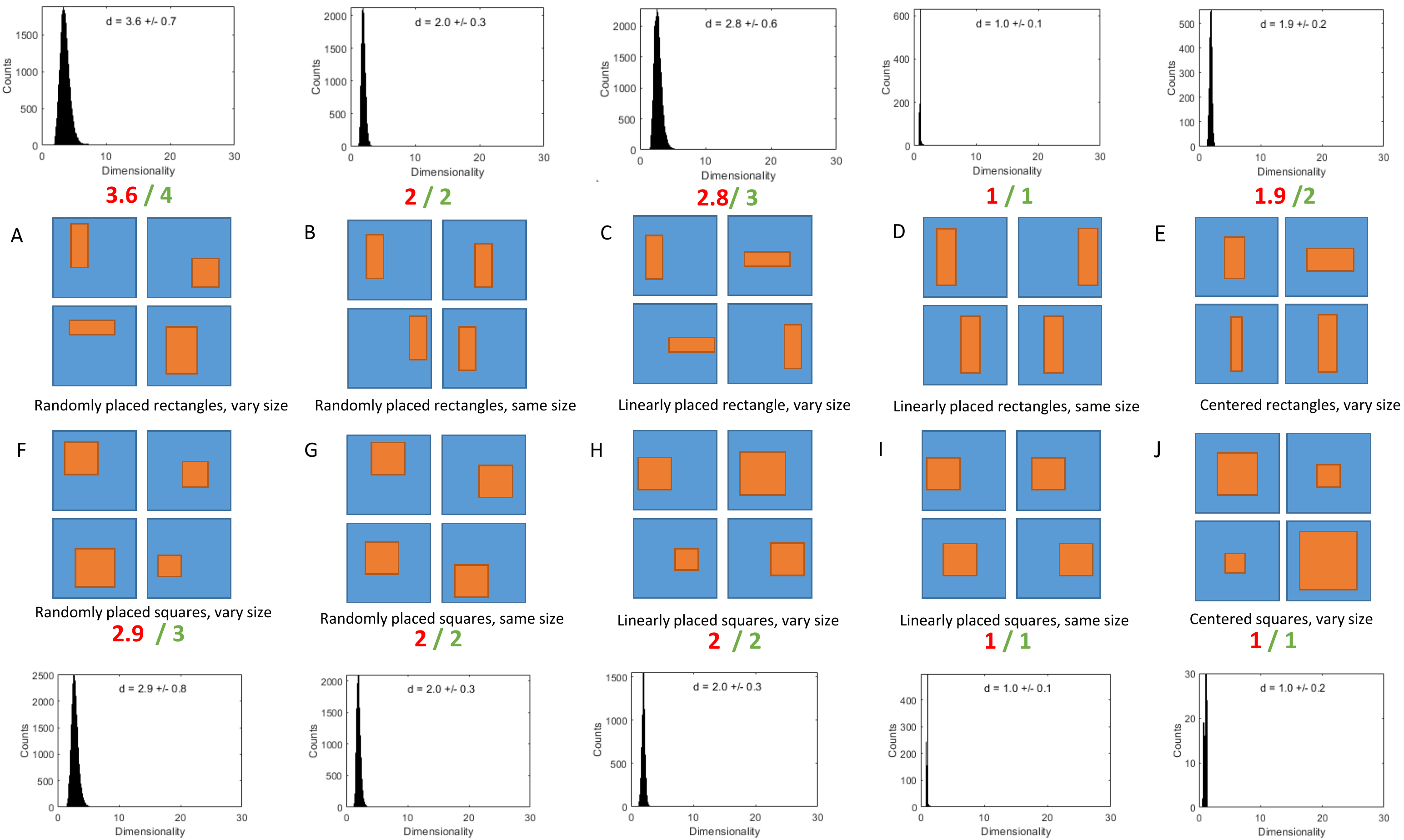}
    \caption{Variation of MLE based intrinsic dimension for a variety of synthetic datasets using $L_1$ norm. The predicted dimension for various data points are shown as a histogram. The numbers in red indicate mean predictions for each case. The numbers in green are the expected dimensionality. }
    \label{figcases}
\end{figure}

To further test the use of $L_1$ norm for MLE estimation of binary images, all ten cases in dataset 1 were tested (results shown in Fig. \ref{figcases}). The intrinsic dimension expected for each case is indicated by numbers in green near the cases in Fig. \ref{figcases}. A histogram of values in the dimension estimate matrix is also plotted and the standard deviation of the histogram is reported in addition to the mean. As seen from these results, the MLE approach with the L1 norm gives a sound estimate  of the intrinsic dimension in all cases. 

\subsection{Retrieving state variables using an autoencoder}

While the MLE algorithm recovers the intrinsic dimensionality, it is of interest to identify the geometry of the latent space and to correlate the dimensions to microstructural features. A variety of applications can benefit from such analysis, including identification of novel processing paths and inverse design of microstructures for a given property as shown in Ref. \cite{sundar2020database}. To generate a proof--of--concept, synthetic dataset C and D are used where the generator space is known. Our objective was to check if the generator space can be retrieved solely from the image data.

\begin{figure}[ht]
    \centering
    \includegraphics[width=0.8\textwidth]{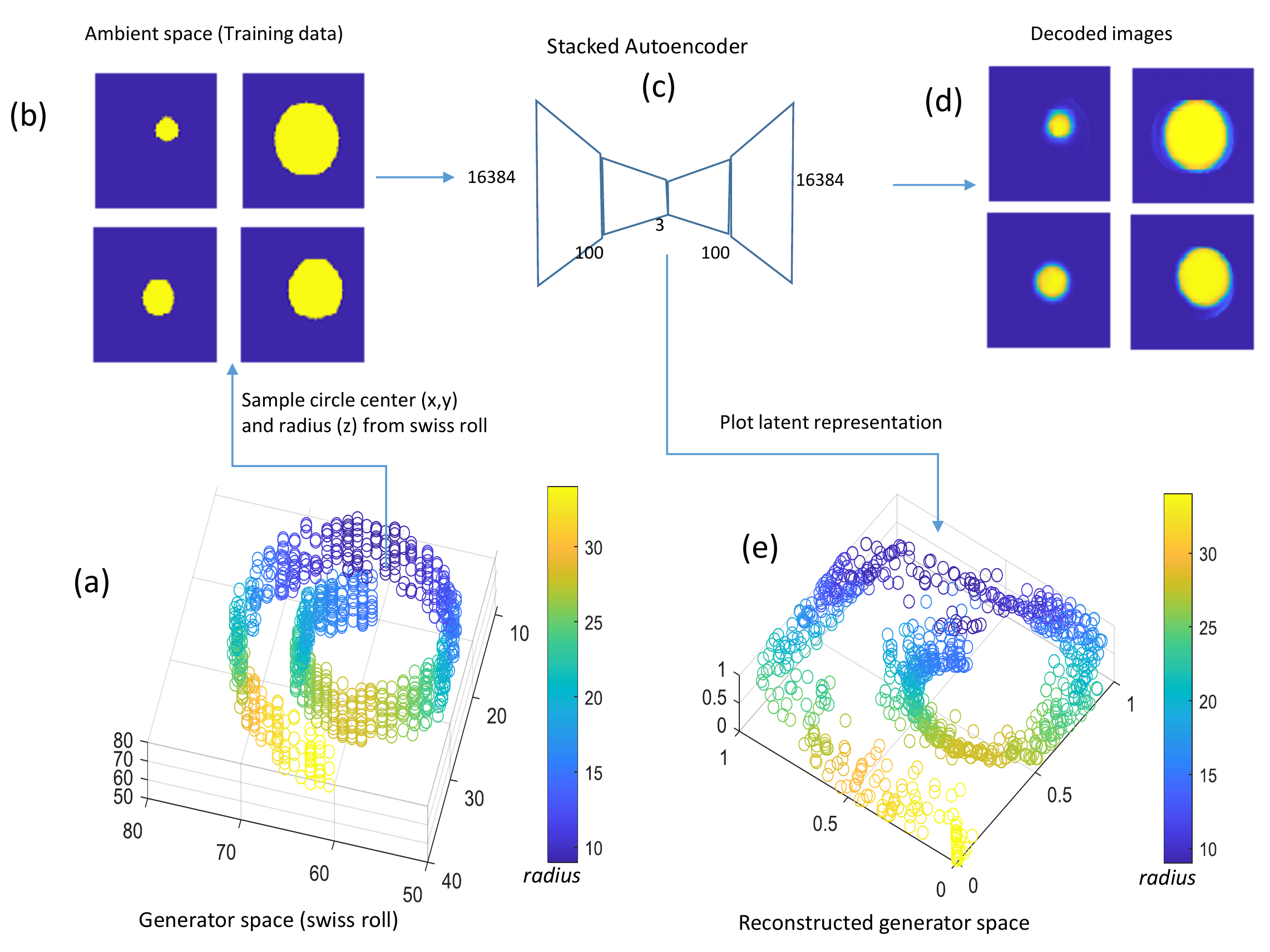}
    \caption{ Use of autoencoders for building a generator space is shown using a synthetic dataset. (a) The swiss roll is the generator space, each point has a $(x_1,x_2,x_3)$ coordinate equal to $(R,x,y)$ in an image where $R$ is the circle radius, $(x,y)$ is the center of the circle. (b) Shows images from the database. (c) Stacked autoencoder reads in the images into a network with a bottleneck equal to the generator space dimension (=3) (d) Decoded images from the last layer of the network. (e) The 1000 images are reduced to three variables in the bottleneck. These variables when plotted show the generator space.}
    \label{figautoencoder}
\end{figure}

The state variables were identified using an autoencoder architecture. An autoencoder (AE)
\cite{baldi1989neural} is a multi-layer neural network that learns the identity function, such that the output $\hat{\mathbf{x}}$ approximates the input $\mathbf{x}$. In the architecture, the hidden layers have fewer nodes than the input dimension and act as a bottleneck. In the first few layers, the autoencoder compresses the input to a compressed (latent space) representation in a process called `encoding'. At its simplest, a single hidden layer operates on the input $\mathbf{x} \in \Re^n$ and generates an encoding $\mathbf{y}_1 \in \Re^j, j < n$ such that:

\begin{equation}
   \mathbf{y}_1 = \sigma(\mathbf{W}_1 \mathbf{x} + \mathbf{b}_1)\label{eq10}
\end{equation}

where $\mathbf{W}_1$ represents the $j \times n$ weight matrix, $\mathbf{b}_1$ is the $j \times 1$ bias  vector for the first layer. The function $\sigma$ is typically a non-linear activation function and a logistic sigmoid function is used in this work. 

Later layers reconstruct the output from this latent space representation in a process called `decoding'. An example is another layer that maps the latent vector $\mathbf{y}_1$ in the previous step to output $\hat{\mathbf{x}} \in \Re^n$ such that 

\begin{equation}
   \hat{\mathbf{x}} = \sigma(\mathbf{W}_2 \mathbf{y}_1 + \mathbf{b}_2)\label{eq11}
\end{equation}
where $\mathbf{W}_2$ represents the $n \times j$ weight matrix, $\mathbf{b}_2$ is the $n \times 1$ bias  vector of the second layer. Multiple layers can be used to develop a deep network. The parameters in $W$ and $b$ are found by minimizing the cost, $ \frac{1}{2}(||\mathbf{x}- \hat{\mathbf{x}}||_2)^2$, by training via backpropagation.\\

In this work, a stacked autoencoder configuration comprised of four layers in total as shown in Fig. \ref{figautoencoder}(c) is employed. The first autoencoder comprised of two layers (encoder and decoder) was trained to reduce the dimensions to 100 first. This was followed by a second autoencoder with two layers (encoder and decoder) that uses the 100 dimensional feature from the first autoencoder as input and reduces it to the intrinsic dimension identified by the MLE algorithm or the generator dimension. The two autoencoders were sequentially trained first, followed by re--training a combined four--layer autoencoder.

Fig. \ref{figautoencoder} shows a schematic of the approach using dataset 3A (circles sampled from a swiss roll, Fig. \ref{figautoencoder}(b)). The stacked autoencoder reads in the images into a network with a bottleneck equal to the generator space dimension (=3). The decoded images from the last layer are shown in Fig. \ref{figautoencoder}(d). The three variables corresponding to each image are plotted in Fig. \ref{figautoencoder}(e) which shows that the generated topology is similar to the actual generator space in Fig. \ref{figautoencoder}(a). The points shown are colored according to the circle radius in the images. Note that the autoencoder, by default, restricts the range of values to between 0 and 1. However, the topology of the space is generally well reconstructed demonstrating a proof--of--concept that variables that define the microstructural state can be identified using stacked autoencoders.

\begin{figure}[ht]
    \centering
    \includegraphics[width=0.8\textwidth]{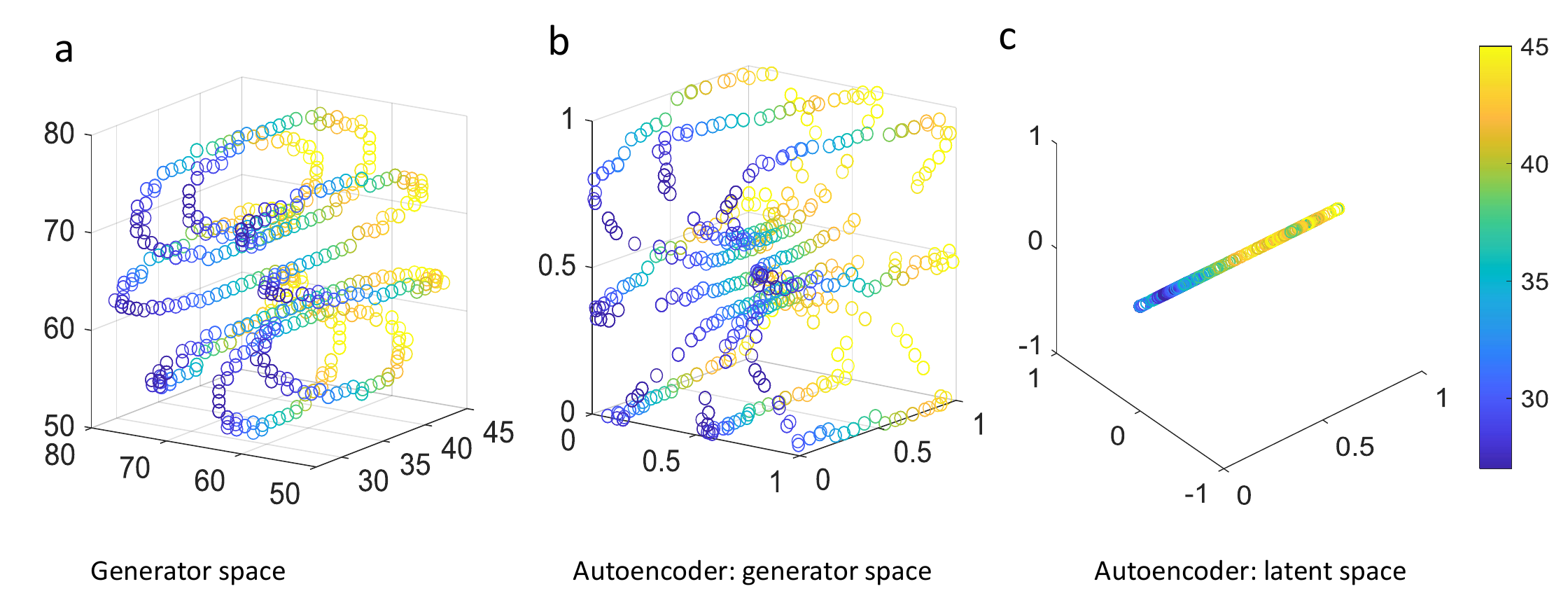}
    \caption{ Helix circles dataset 3B: (a) Generator space where images are sampled (b) Generator space computed from the images using the stacked autoencoder (c) Intrinsic dimensional (1D) space computed by the stacked autoencoder.}
    \label{figlatent}
\end{figure}

Figure \ref{figlatent}(b,c) shows both the computed generator space and the 1D latent space for the helix circles dataset 3B (Figure \ref{figlatent}(a) shows the actual generator space from which the images were sampled with $(x_1,x_2,x_3)$ coordinate equal to $(r,x,y)$ in the image). As in the swiss roll case, the generator space for the helix data set also looks similar to the generator space considering that the points are mapped in the range $[0,1]$ by the autoencoder. The computed latent spaces are colored according to a microstructural feature, the radius of the circle. Clustering of this feature in the latent space demonstrates promise towards the use of proximity analysis to find new microstructures with interesting properties through interpolation. Such an application will form a part of our future efforts.

\begin{figure}[ht]
    \centering
    \includegraphics[width=1.0\textwidth]{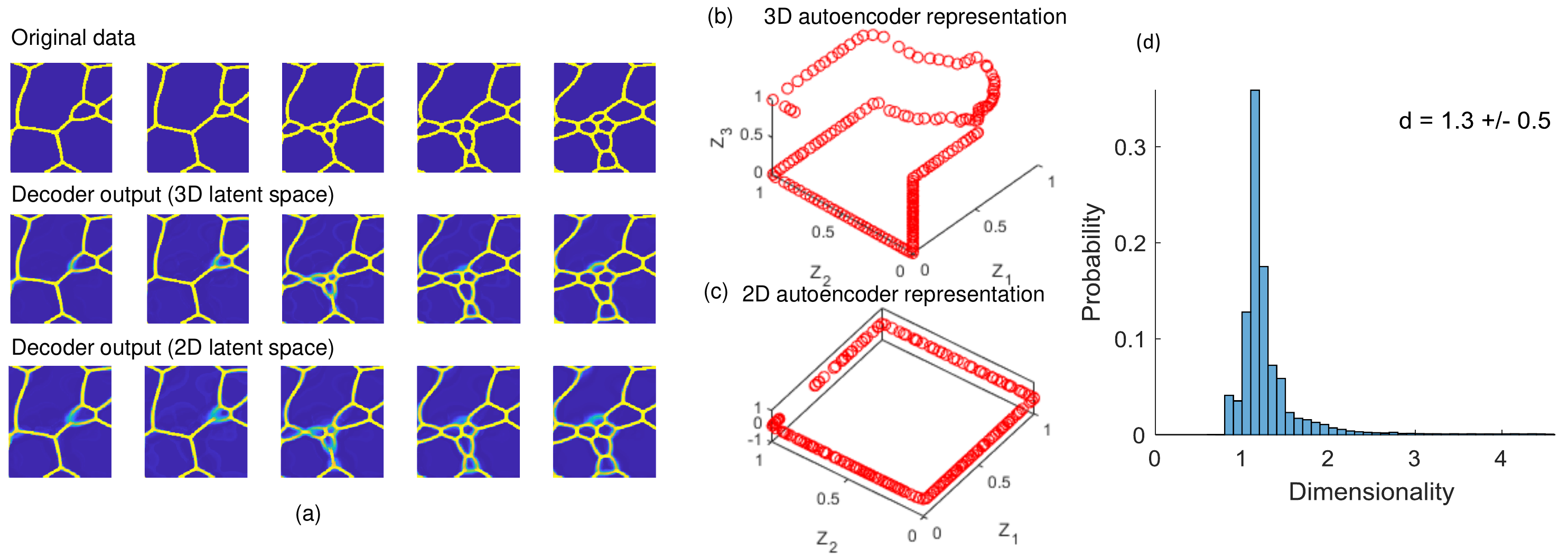}
    \caption{(a) Decoder outputs for a single phase field trajectory (b) Comparison of 3D versus 2D representation of the autoencoder (c) MLE estimate of dimensionality}
    \label{pf1}
\end{figure}

The last example is from dataset 4 (phase field data). In the results from this dataset in Fig. \ref{pf1},  one simulation trajectory was used in two stacked autoencoders of bottleneck 3 and 2. Fig. \ref{pf1}(a) compares the reconstructions from the autoencoder against the original images at five randomly chosen time steps from this trajectory. Each consecutive time step in the phase field data results in incremental changes in topology of grain boundaries, hence it is expected that data points are arranged in the order of time steps in the latent space. This is indeed seen in the topologies of the space constructed by the autoencoder as shown in Fig. \ref{pf1}(b,c). Similar to the helix case seen earlier, the intrinsic dimensionality is expected to be one which is confirmed by the results of the MLE algorithm in Fig. \ref{pf1}(c). 

\begin{figure}[ht]
    \centering
    \includegraphics[width=1.0\textwidth]{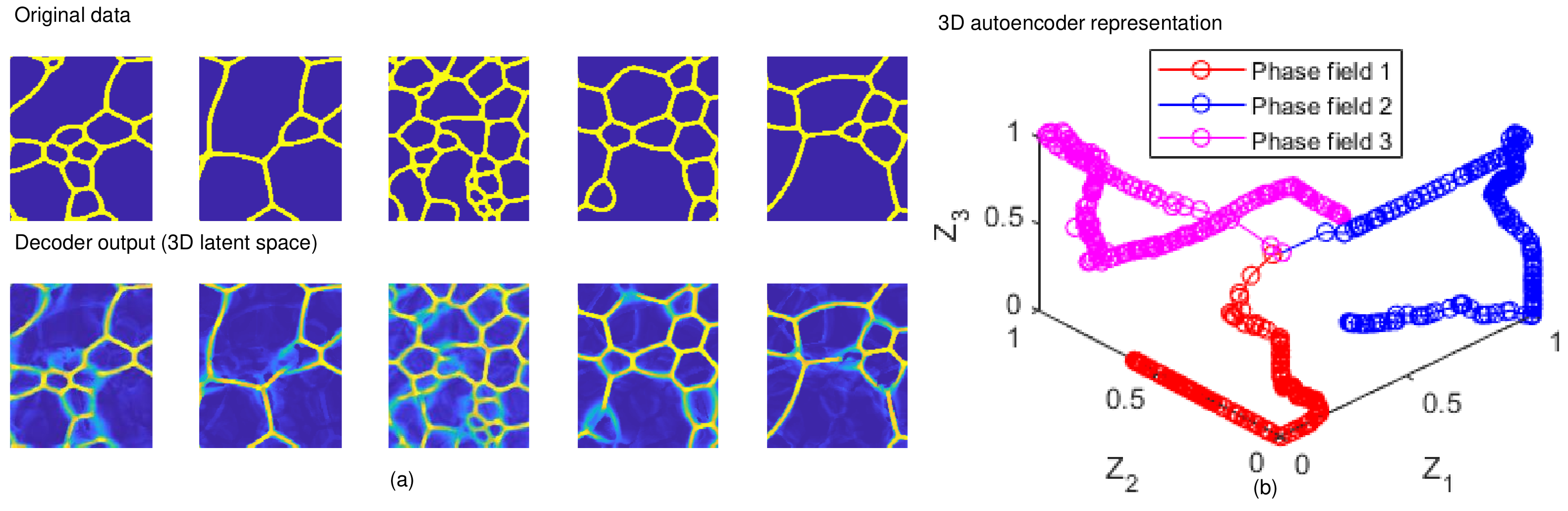}
    \caption{(a) Decoder outputs for combined dataset containing all three phase field trajectories  (b) 3D representation from autoencoder showing all three trajectories.}
    \label{pf2}
\end{figure}

In the last example, the complete dataset containing three different phase field simulations is employed with an autoencoder bottleneck of 3. The resulting latent space is shown in Fig. \ref{pf2}. In all three simulations, the initial image was the same represented by the central point in the latent space. The trajectories from the three simulations emerge in different directions from the initial point, resulting in different final microstructures. This latent space is an example of a microstructural space for a grain coarsening process. To estimate the true dimensionality of the processing space, a large number of trajectories need to be superposed. Although a simplistic set of three trajectories are shown, this example shows how the framework can be used to visualize a multitude of complex processes within a single graph. Past work in \cite{mssl8,mssl38} employed similar visualization of linear PCA components to perform process design, the use of non--linear manifolds as demonstrated here is expected to significantly improve state--of--the--art and will form part of our future work.

\section{Conclusions}

A methodology for reliably estimating the intrinsic dimensionality of random media is developed. The method resolves the ambiguity in results for images with low bit depth when using state--of--the--art techniques that employ the Euclidean ($L_2$) norm. Particular novel contributions of this work are listed below:

\begin{itemize}
    \item It is shown that the NN and MLE formulae for intrinsic dimensionality estimation can work with all Minkowski distance measures. Further, the examples show that all Minkowski measures give the same intrinsic dimensionality for non-discrete datasets.
    \item Dimensionality estimates are dependent on distance measures for image data with discrete levels as was shown for binary images. Through examples, it is shown that the use of $L_1$ distance in the MLE estimate produces a reliable estimate in such cases.
    \item Examples show that stacked autoencoders can reconstruct the low dimensional generator spaces of microstructures and provide a sparse set of state variables to fully describe material microstructures. 
\end{itemize}
Use of these state variables to quantify microstructure--property--process relationships will form a part of our future work.

\section*{Data availability}
Codes and datasets will be made freely available upon publication in a peer--reviewed journal.

\section*{Acknowledgements}
This research was supported in part by the Air Force Research Laboratory Materials and Manufacturing Directorate, through the Air Force Office of Scientific Research Summer Faculty Fellowship Program, Contract Numbers FA8750-15-3-6003 and FA9550-15-0001. This research was supported in part through computational resources and services provided by Advanced Research Computing at the University of Michigan, Ann Arbor.

\bibliographystyle{plain}
\bibliography{References}

\section*{Appendix 1} 
\setcounter{figure}{0} 
\makeatletter 
\renewcommand{\fnum@figure}{Figure S\thefigure} 

In the datasets used in this paper, we avoided the intersection of the particle shape with the image boundary to get an unbiased estimation of the dimensionality. In general, image boundaries can play a role in biasing the intrinsic dimension. Consider the case of a single circular shape placed in a matrix. If the shape intersects the boundary, only a part of the shape is seen and the intrinsic dimensionality of three as judged from the dataset assumes that the shape always remains a circle. To test this, we have plotted results from two datasets, one with and one without boundary intersections. We consider sufficiently high sample size ($3000$ images) and image size ($128\times128$) for each case. 

\begin{figure}[ht]
    \centering
    \includegraphics[width=1.0\textwidth]{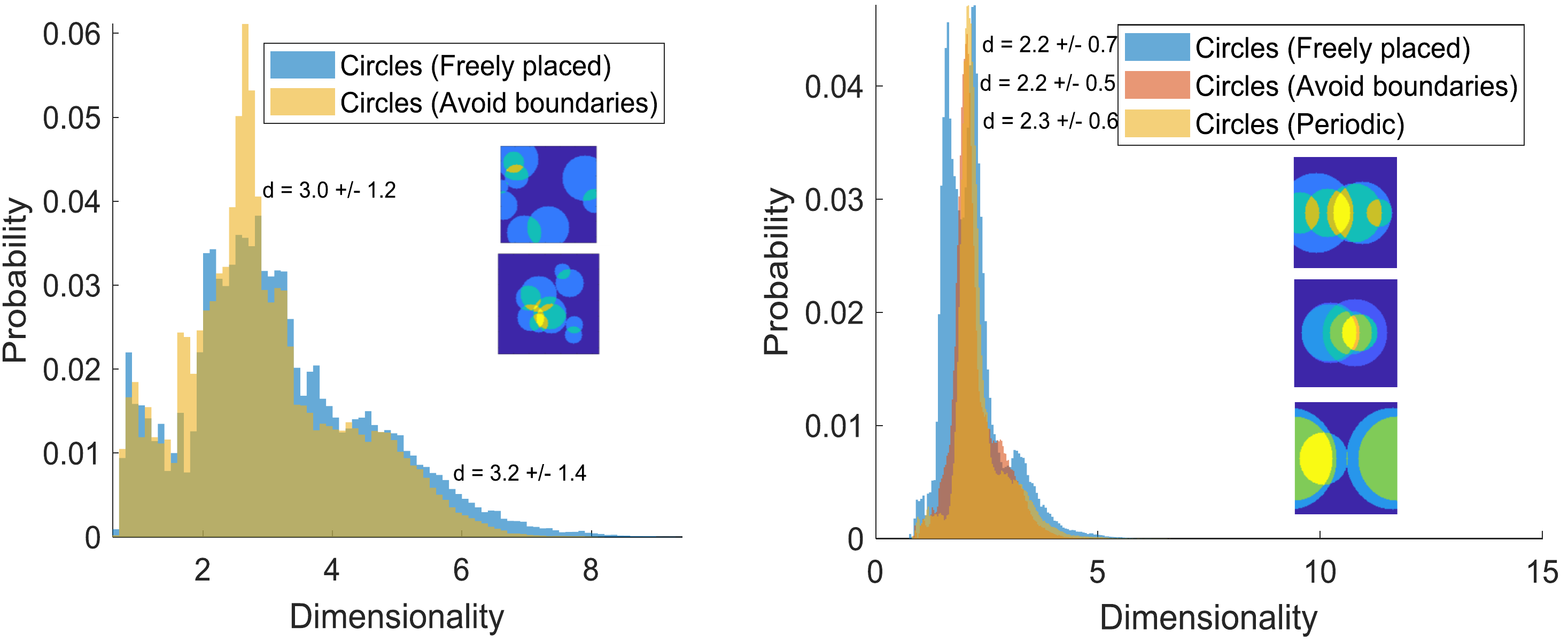}
    \caption{Histograms of estimated dimension per data point (a) Two cases are considered, a circle of varying radii is freely placed in one case and avoids boundary in another case. (b) Here, the circle is linearly placed along the centerline. A case with periodicity is included. Inset images show superposition of a few different images in each dataset.}
    \label{figboundary}
\end{figure}

The histogram of estimated dimension per data point is plotted in Fig. S\ref{figboundary}a showing that the algorithm is able to predict the correct mean dimension for both cases. The key difference being that the histogram is broader and a higher standard deviation is obtained when the circles intersect the boundary. Another case is shown in Fig. S\ref{figboundary}b where the circle is linearly placed along the centerline. Here, two distinct peaks are seen in the histogram where the circles are freely placed, with the first peak at a lower intrinsic dimension. the mean dimension is again correctly estimated for both cases with a higher standard deviation for the case where circles intersect the image boundaries. A case where the circles are periodic is also shown here, where the circles wrap around on the opposite side when they intersect the boundary. While this case shows a single sharp peak as in the case where boundaries are avoided, the standard deviation is higher than that case.

\section*{Appendix 2} 

Equation (10) simplifies to:

\begin	{equation}
\begin	{split}
F_k(r_p) &= \left(\frac{(c(p)r_p^\mu)^{k-1}}{\Gamma(k)}e^{-c(p)r_p^\mu}\right)c(p)\mu r_p^{\mu -1}\\
&=\left[c(p)^{k}\mu\left(\frac{1}{\Gamma(k)}\right)\right]r_p^{\mu k -1}e^{-c(p)r_p^\mu}\\
\end	{split} \nonumber
\end	{equation}

Inserting this into Equation (11):
\nopagebreak
\begin{equation}
\begin	{split}
E_k(r_p) &= \int_0^\infty r_p F_k(r_p) dr_p\\
&= \left[c(p)^k\mu\left(\frac{1}{\Gamma(k)}\right)\right]\int_0^\infty r_p^{\mu k}e^{-c(p)r_p^\mu} dr_p\\
\end	{split}  \nonumber
\end{equation}

Changing the integration variable with $\xi \triangleq c(p)r_p^\mu$ ($r_p = \xi^{1/\mu}c(p)^{-1/\mu}$):
\nopagebreak
\begin{equation}
\begin	{split}
E_k(r_p) &= \left[c(p)^k\mu\left(\frac{1}{\Gamma(k)}\right)\right]\int_0^\infty (\xi^{1/\mu}c(p)^{-1/\mu})^{\mu k}e^{-\xi} d(\xi^{1/\mu}c(p)^{-1/\mu})\\
&= c(p)^{-1/\mu}\left(\frac{1}{\Gamma(k)}\right)\int_0^\infty \xi^{k + 1/\mu -1}e^{-\xi} d\xi\\
\end	{split}  \nonumber
\end{equation}

The integral is, by definition, $\Gamma(k + 1/\mu)$ ~\cite{gradshteyn2014table}. Substituting:
\nopagebreak
\begin{equation}
E_k(r_p) = c(p)^{-1/\mu}\frac{\Gamma(k + 1/\mu)}{\Gamma(k)}  \nonumber
\end{equation}

\end{document}